\pdfoutput=1
\documentclass[11pt,a4paper]{article}
\usepackage[hyperref]{emnlp2018}
\usepackage{times}
\usepackage{latexsym}
\usepackage{amsmath}
\usepackage{amsthm}
\usepackage{latexsym}
\usepackage{array}
\usepackage{ccaption} 
\usepackage{wrapfig}
\usepackage{multirow}
\usepackage{setspace}
\usepackage{colordvi}  
\usepackage{color}
\usepackage{graphicx}
\usepackage{epsfig}
\usepackage{epstopdf}
\usepackage{relsize}
\usepackage{url}

\aclfinalcopy % Uncomment this line for the final submission

\title{BLEU is Not Suitable for the Evaluation of Text Simplification}

\author{Elior Sulem, Omri Abend, Ari Rappoport \\
 Department of Computer Science, The Hebrew University of Jerusalem\\
{\tt \{eliors|oabend|arir\}@cs.huji.ac.il}
 }

\date{}

\begin{document}
\maketitle
\begin{abstract}
  
  BLEU is widely considered to be an informative
  metric for text-to-text generation, including Text Simplification (TS).
  TS includes both lexical and structural aspects.
  In this paper we show that BLEU is not suitable for the evaluation of sentence splitting, the major
  structural simplification operation. We manually compiled a sentence splitting gold standard corpus
  containing multiple structural paraphrases, and performed a correlation analysis with human judgments.\footnote{The corpus can be found in \url{https://github.com/eliorsulem/HSplit-corpus}}
  We find low or no correlation between BLEU and the grammaticality and meaning preservation parameters
  where sentence splitting is involved. Moreover, BLEU often negatively correlates with
  simplicity, essentially penalizing simpler sentences.
\end{abstract}

%%%%%%%%%%%%%%%%%%%%%%%%%%%%%%%%%%%%%%%%%%%%%%%%%%%%%%%%%%%%%%%%
\section{Introduction} \label{sec:introduction}

BLEU \citep{P02} is an $n$-gram-based evaluation metric, widely used for Machine Translation (MT) evaluation. BLEU has also been 
applied to monolingual translation tasks, such as grammatical error correction \citep{PL11}, summarization \citep{G15} and text simplification \citep{NG14,Sa15,Xu16}, i.e.
the rewriting of a sentence as one or more simpler sentences.

Along with the application of parallel corpora and MT techniques for TS \citep[e.g.,][]{Z10,W12,NG14},
BLEU became the main automatic metric for TS,
despite its deficiencies (see \S\ref{sec:related_work}).
Indeed, focusing on lexical simplification, \citet{Xu16} argued that BLEU
gives high scores to sentences that are close or even identical to the input,
especially when multiple references are used.
In their experiments, BLEU failed to predict simplicity,
but obtained a higher correlation with grammaticality and meaning preservation,
relative to the SARI metric they proposed.

In this paper, we further explore the applicability of BLEU for TS evaluation, examining BLEU's
informativeness where sentence splitting is involved.
Sentence splitting, namely the rewriting of a single sentence as multiple sentences while preserving its meaning, is the main structural simplification operation. It has been shown useful for MT preprocessing \citep{C96,M14,LN15} and human comprehension \citep{MK79,W03}, independently from other lexical and structural simplification operations.
Sentence splitting is performed by many TS systems \citep{Z10,WL11,SA14,NG14,NG16}.
For example, 63$\%$ and 80$\%$ of the test sentences are split by the systems of \citet{WL11} and \citet{Z10}, respectively \citep{NG16}.
Sentence splitting is also the focus of the recently proposed Split-and Rephrase sub-task \citep{N17,AG18}, in which the automatic metric used is BLEU.

For exploring the effect of sentence splitting on BLEU scores, we compile a human-generated
gold standard sentence splitting corpus -- HSplit,
which will also be useful for future studies of splitting in TS,
and perform correlation analyses with human judgments.
We consider two reference sets. First, we experiment with the most common set, proposed by \citet{Xu16}, evaluating
a variety of system outputs, as well as HSplit.
The references in this setting explicitly emphasize lexical operations, 
and do not contain splitting or content deletion.\footnote{Nevertheless, they are also used in contexts where structural operations are involved \citep{Ni17,S18acl}.}
Second, we experiment with HSplit as the reference set, evaluating systems that focus on sentence splitting.
The first setting allows assessing whether BLEU with the standard reference set is a reliable metric on systems that perform splitting.
The second allows assessing whether BLEU can be adapted to evaluate splitting, given a reference set so oriented.

We find that BLEU is often negatively correlated with simplicity, even
when evaluating outputs without splitting, and that when evaluating outputs with splitting,
it is less reliable than a simple measure of similarity to the source (\S\ref{sec:results}).
Moreover, we show that BLEU cannot be adapted to assess sentence splitting,
even where the reference set focuses on this operation (\S\ref{sec:splitting_refs}).
We conclude that BLEU is not informative and is often misleading for TS evaluation and for the related Split and Rephrase task.

\section{Related Work} \label{sec:related_work}

\paragraph{The BLEU Metric.}

BLEU \citep{P02} is reference-based, where the use of multiple references is used to address cross-reference variation.
To address changes in word order, BLEU uses $n$-gram precision, 
modified to eliminate repetitions across the references. A brevity term penalizes overly short sentences. Formally:

\vspace{-0.2cm}
\begin{small}
 $${\rm BLEU} = {\rm BP} \times {\rm exp}(\sum_{n=1}^{N} w_{n} {\rm log}(p_{n}))$$
\end{small}
\vspace{-0.3cm}

where ${\rm BP}$ is the brevity penalty term, $p_n$ are the modified precisions, and $w_n$ are the corresponding weights,
which are usually uniform in practice.

The experiments of \citet{P02} showed that BLEU correlates with human judgments in the ranking of five English-to-Chinese MT systems and that it can distinguish human and machine translations. Although BLEU is widely used in MT, several works have pointed out its shortcomings \citep[e.g.,][]{KM06}. In particular, \citet{C06} showed that BLEU may not correlate in some cases with human judgments since a huge number of potential translations have the same BLEU score, and that correlation decreases when translation quality is low. Some of the reported shortcomings are relevant to monolingual translation, such as the impossibility to capture synonyms and paraphrases that are not in the reference set, or the uniform weighting of words. 

\paragraph{BLEU in TS.}
While BLEU is standardly used for TS evaluation \citep[e.g.,][]{Xu16,Ni17,ZL17,MS17}, only few works tested its correlation with human judgments. Using 20 source sentences from the PWKP test corpus \citep{Z10} with 5 simplified sentences for each of them, \citet{W12} reported positive correlation of BLEU with simplicity ratings, but no correlation with adequacy.
T-BLEU \citep{S14}, a variant of BLEU which uses lower n-grams when no overlapping 4-grams are found,
was tested on outputs that applied only structural modifications to the source.
It was found to have moderate positive correlation for meaning preservation, and positive but low correlation
for grammaticality. Correlation with simplicity was not considered in this experiment.
\citet{Xu16} focused on lexical simplification, finding that BLEU obtains reasonable correlation for grammaticality
and meaning preservation but fails to capture simplicity, even when multiple references are used.
To our knowledge, no previous work has examined the behavior of BLEU on sentence splitting, which we investigate here using a manually compiled gold standard.

\section{Gold-Standard Splitting Corpus} \label{sec:corpus}

In order to investigate the effect of correctly splitting sentences on the automatic metric scores, we build a parallel corpus, 
where each sentence is modified by 4 annotators, according to specific sentence splitting guidelines.
We use the complex side of the test corpus of \citet{Xu16}.\footnote{\url{https://github.com/cocoxu/simplification} includes the corpus, the SARI metric and the SBMT-SARI system. The corpus comprises 359 sentences.} 

While \citet{N17} recently proposed the semi-automatically compiled WEB-SPLIT dataset for training automatic sentence splitting systems, here we generate a completely manual corpus, without a-priori splitting points nor do we pre-suppose that all sentences should be split. This corpus enriches the set of references focused on lexical operations that were collected by \citet{Xu16} for the same source sentences and can also be used as an out-of-domain test set for Split-and-Rephrase \citep{N17}.

We use two sets of guidelines.
In Set 1, annotators are required to split the original as much as possible,
while preserving the sentence's grammaticality, fluency and meaning.
The guidelines include two sentence splitting examples.\footnote{Examples are taken from \citet{S06}.}
In Set 2, annotators are encouraged to split only in cases where it simplifies the original sentence.
That is, simplicity is implicit in Set 1 and explicit in Set 2. In both sets, the annotators are instructed to leave
the source unchanged if splitting violates grammaticality,
fluency or meaning preservation.\footnote{Examples are not provided in the case of Set 2 so as not to give an a-priori notion of simplicity. The complete guidelines are found in the supplementary material.} 

Each set of guidelines is used by two annotators, with native or native-like proficiency in English.
The obtained corpora are denoted by HSplit1, HSplit2 (for Set 1), and HSplit3 and HSplit4 (for Set 2), each containing 359 sentences. Table \ref{tab:corpus_stats} presents statistics for the corpora. Both in terms of the number of splits per sentence ($\#$ Sents) and in terms of the proportion of input sentences that have been split (SplitSents), we observe that the average difference within each set is significantly greater than the average difference between the sets.\footnote{Wilicoxon's signed rank test, $p = 1.6 \cdot 10^{-5}$ for $\#$Sents and $p = 0.002$ for SplitSents.}
This suggests that the number of splits is less affected by the explicit mention of simplicity than by the inter-annotator variability. 

\begin{center}
\begin{table}[h]
\scriptsize
\begin{center}
\begin{tabular}{|c|c|c|}
\hline
& {\bf $\#$ Sents} & {\bf SplitSents ($\%$)}\\
\hline
{\bf HSplit1} & 1.93 & 68\\
\hline
{\bf HSplit2 } & 2.28&  86\\
\hline  
{\bf HSplit3} & 1.87& 63\\
\hline
{\bf HSplit4} & 1.99& 71\\
\hline
{\bf HSplitAverage} &2.02  & 72\\
\hline    
\end{tabular}
\end{center}
\hfill
\caption{\small\label{tab:corpus_stats}Statistics for the sentence splitting benchmark. $\#$Sents denotes the average number of sentences in the output. SplitSents denotes the proportion of input sentences that have been split. 
The last row presents the average scores of the 4 HSplit corpora.}
\label{tab:split_effect}
\end{table}
\end{center}

\begin{center}
\begin{table*}[ht]
\scriptsize
\centering
\begin{tabular}{|c|c|c|c|c||c|c|c|c|}
\hline
& \multicolumn{4}{|c||}{{\bf Systems/Corpora without Splits}} & \multicolumn{4}{|c|}{{\bf All Systems/Corpora}}\\
\hline
&{\bf G} & {\bf M} & {\bf S} & {\bf StS} &{\bf G} & {\bf M} & {\bf S} & {\bf StS} \\
\hline
{\bf BLEU-1ref}& 0.43 (0.2) & 1.00 (0) & -0.81 (0.01) & -0.43 (0.2) & 0.11 (0.4) & 0.08 (0.4) & -0.60 (0.02) & -0.67 (0.008)  \\
\hline   
{\bf BLEU-8ref} & 0.61 (0.07) & 0.89 (0.003) & -0.59 (0.08) & -0.11 (0.4) & 0.26 (0.2) & 0.13 (0.3) & -0.42 (0.08) & -0.50 (0.05)\\
\hline
{\bf iBLEU-1ref} & 0.21 (0.3) & 0.93 (0.001) & -0.85 (0.008) & -0.61 (0.07) & 0.02 (0.5) & 0.07 (0.4) & -0.61 (0.02) & -0.71(0.004) \\
\hline     
{\bf iBLEU-8ref} &0.61 (0.07)& 0.89 (0.003) & -0.59 (0.08) & -0.11 (0.4) & 0.26 (0.2) & 0.13 (0.3) & -0.42 (0.08)& -0.50 (0.05) \\
\hline
{\bf -FK} & -0.21 (0.3) & -0.57 (0.09) & 0.67 (0.05)& 0.39 (0.2) & -0.05 (0.4) & -0.03 (0.5) & 0.51 (0.05)& 0.64 (0.01) \\
\hline
 {\bf SARI-8ref} & -0.64 (0.06)& -0.86 (0.007) & 0.52 (0.1)& 0.00 (0.5) & -0.64 (0.01)& -0.72 (0.004) & 0.26 (0.2)& -0.02 (0.5)\\
\hline \hline
{\bf -LD$_{SC}$} & 0.29 (0.3)& 0.86 (0.007)  & -0.88 (0.004) & -0.57 (0.09)& 0.21 (0.3)& 0.51 (0.04)& -0.68 (0.007) & -0.52 (0.04)\\
\hline   
\end{tabular}
\hfill
\caption{\small Spearman correlation (and $p$-values) at the system level between the rankings of automatic metrics and of human judgments for ``Standard Reference Setting''.
Automatic metrics (rows) include BLEU and iBLEU (each used either with a single reference or with 8 references), the negative Flesh-Kincaid Grade Level (-FK), and SARI, computed with 8 references. We also include the negative Levenshtein distance between the output and the source (-LD$_{SC}$). Human judgments are of the Grammaticality (G), Meaning Preservation (M), Simplicity (S) and Structural Simplicity (StS) of the output.
  The left-hand side reports correlations where only simplifications that do not include sentence splitting are considered. The right-hand side reports correlations where the HSplit 
  corpora are evaluated as well (see text).
  BLEU negatively correlates with S and StS in both cases, and shows little to no correlation with G and M where sentence splitting is involved.
}
\label{tab:correlation}
\end{table*}
\end{center}

\vspace{-1cm}
\section{Experiments}\label{sec:correlation}

\subsection{Experimental Setup}\label{sec:systems}

\paragraph{Metrics.} \label{sec:metrics}
In addition to BLEU,\footnote{System-level BLEU scores are computed using the multi-bleu Moses support tool. Sentence-level BLEU scores are computed using NLTK \citep{LB02}.} we also experiment with (1) iBLEU \citep{SZ12} which was recently used for TS \citep{Xu16,Z17} and
which takes into account the BLEU scores of the output against the input and against the references;
(2) the Flesch-Kincaid Grade Level \citep[FK;][]{K75}, computed at the system level, which estimates the readability of the text with a lower value
indicating higher readability;\footnote{We thus computed the correlation in \S\ref{sec:results} for -FK.} 
(3) SARI \citep{Xu16}, which compares the $n$-grams of the system output with those of the input and the human references, separately evaluating the quality of words that are added, deleted and kept by the systems. 
For completeness, we also experiment with the negative Levenshtein distance to the source (-LD$_{SC}$), which serves as a measure of conservatism.\footnote{LD$_{SC}$ is computed using NLTK.}

We explore two settings. In one (``Standard Reference Setting'', \S\ref{sec:results}), we use two sets of references: the Simple Wikipedia reference (yielding BLEU-1ref and iBLEU-1ref), and 8 references obtained by crowdsourcing by \citet{Xu16} (yielding BLEU-8ref, iBLEU-8ref and SARI-8ref). In the other (``HSplit as Reference Setting'', \S\ref{sec:splitting_refs}), we use HSplit as the reference set.

\paragraph{Systems.} \label{sec:systems}
For ``Standard Reference Setting'', we consider both a case where evaluated systems do not perform any splittings on the test set (``Systems/Corpora without Splits''), 
and one where we evaluate these systems, along with the HSplit corpus, used in the role of system outputs (``All Systems/Corpora'').
Systems include six MT-based simplification systems, including outputs of the state-of-the-art neural TS system of \citet{Ni17}, 
 in four variants: either default settings or initialization by word2vec, for each both the highest and the fourth ranked hypotheses in the beam are considered.\footnote{Taking the fourth hypothesis rather than the first has been found to yield considerably less conservative TS systems.}
We further include Moses \citep{K07}
and SBMT-SARI \citep{Xu16}, a syntax-based MT system tuned against SARI, and the identity function (outputs are same as inputs). 
The case which evaluates outputs with sentence splitting additionally includes the four HSplit corpora and the HSplit average scores.

For ``HSplit as Reference Setting'', we consider the outputs of six simplification systems whose main simplification operation is sentence splitting: DSS, DSS$^{m}$, SEMoses, SEMoses$^{m}$, SEMoses$_{LM}$ and SEMoses$_{LM}^m$, taken from \citep{S18acl}. 

\paragraph{Human Evaluation.}\label{sec:human_evaluation}
We use the evaluation benchmark provided by \citet{S18acl},\footnote{\url{https://github.com/eliorsulem/simplification-acl2018}} including system outputs and human evaluation scores corresponding to the first 70 sentences of the test corpus of \citet{Xu16}, and extend it to apply to HSplit as well.

The evaluation of HSplit is carried out by 3 in-house native English annotators, 
who rated the different input-output pairs for the different systems according to 4 parameters: Grammaticality (G), 
Meaning preservation (M), Simplicity (S) and Structural Simplicity (StS). 
G and M are measured using a 1 to 5 scale. A -2 to +2 scale is used for measuring simplicity and structural simplicity.
For computing the inter-annotator agreement of the whole benchmark (including the system outputs and the HSplit corpora), we follow \citet{PT16} and randomly select, for each sentence, one annotator's rating to be the rating of Annotator 1 and the rounded average rating of the two other annotators to be the rating of Annotator 2. We then compute weighted quadratic $\kappa$ \citep{C68} between Annotator 1 and 2. Repeating this process 1000 times, the obtained medians and 95$\%$ confidence intervals are 0.42 $\pm$ 0.002 for G, 0.77 $\pm$ 0.001 for M and 0.59 $\pm$ 0.002 for S and StS.

\subsection{Results with Standard Reference Setting} \label{sec:results}

\paragraph{Description of the Human Evaluation Scores.}

The human evaluation scores for each parameter are obtained by averaging over the 3 annotators. The scores at the system level are obtained by averaging over the 70 sentences. In the "All systems/corpora" case of the "Standard Reference Setting", where 12 systems/corpora are considered, the range of the average G scores at the system level is from 3.71 to 4.80 ($\sigma$ = 0.29). For M, this max-min difference between the systems is 1.23 ($\sigma$=0.40). For S and StS, the differences are 0.53 ($\sigma$ = 0.17) and 0.65 ($\sigma$ = 0.20). At the sentence level, considering 840 sentences (70 for each of the system/corpora), the G and M scores vary from 1 to 5 ($\sigma$ equals 0.69 and 0.85 respectively), and the S and StS scores from -1 to 2 ($\sigma$ equals 0.53 and 0.50).

In the "Systems/corpora without Splits" case of the "Standard Reference Setting", where 7 systems/corpora are considered, the max-min difference at the system level are again  1.09  ($\sigma$ = 0.36) and 1.23 ($\sigma$ = 0.47) for G and M respectively. For S and StS, the differences are 0.45 and 0.49 ($\sigma$ = 0.18). At the sentence level, considering 490 sentences (70 for each of the system/corpora), the G and M scores vary from 1 to 5 ($\sigma$ equals 0.78 and 1.01 respectively), and the S and StS scores from -1 to 2 ($\sigma$ equals 0.51 and 0.46).

\paragraph{Comparing HSplit to Identity.}

Comparing the BLEU score on the input (the identity function) and on the HSplit corpora, we observe that the former yields much higher BLEU scores. Indeed, BLEU-1ref obtains 59.85 for the input and 43.90 for the HSplit corpora (averaged over the 4 HSplit corpora). BLEU-8ref obtains 94.63 for the input and 73.03 for HSplit.\footnote{These scores concern the first 70 sentences of the corpus. A similar phenomenon is observed on the whole corpus (359 sentences). BLEU-1ref obtains 59.23 for the input and 45.68
for HSplit. BLEU-8ref obtains 94.93 for the input and 75.68 for HSplit.} The high scores obtained for Identity, also observed by \citet{Xu16}, indicate that BLEU is a not a good predictor for relative simplicity to the input. The drop in the BLEU scores for HSplit is not reflected by the human evaluation scores for grammaticality (4.43 for AvgHSplit vs. 4.80 for Identity) and meaning preservation (4.70 vs. 5.00), where the decrease between Identity and HSplit is much more limited. For examining these tendencies in more detail, we compute the correlations between the automatic metrics and the human evaluation scores. They are described in the following paragraph.
  
\paragraph{Correlation with Human Evaluation.}
The system-level Spearman correlations between the rankings of the automatic metrics and the human judgments
(see \S\ref{sec:systems}) are presented in Table \ref{tab:correlation}.
We find that in all cases BLEU and iBLEU negatively correlate with S and StS, indicating that they fail to capture simplicity and structural simplicity.
Where gold standard splits are evaluated as well, 
BLEU's and iBLEU's failure to capture StS is even more pronounced.
Moreover, BLEU's correlation with G and M in this case disappears. In fact,
BLEU's correlation with M in this case
is considerably lower than that of -LD$_{SC}$ and its correlation with G is comparable, suggesting BLEU is 
inadequate even as a measure of G and M if splitting is involved.

We examine the possibility that BLEU mostly acts as a measure of conservatism,
and compute the Spearman correlation between -LD$_{SC}$ and BLEU.
The high correlations we obtain between the metrics indicate that this may be the case.
Specifically, BLEU-1ref obtains correlations of 0.86 ($p= 7 \times 10^{-3}$) without splits and of 0.52 ($p=0.04$) where splitting is involved.
BLEU-8ref obtains 
0.82 ($p=0.01$) and 0.55 ($p=0.03$).

SARI obtains positive correlations with S, of 0.52 (without splits) and 0.26 (all systems/corpora),
but correlates with StS in neither setting.
This may stem from SARI's focus on 
lexical, rather than structural TS. 

Similar trends are observed in the sentence-level correlation for S, StS and M, whereas G sometimes benefits in the sentence level from including HSplit in the evaluation. For G and M, the correlation with BLEU is lower than its correlation with -LD$_{SC}$ in both cases.

\begin{center}
\begin{table}[h]
\scriptsize
\centering
\setlength\tabcolsep{4pt}
\begin{tabular}{|c|c|c|c|c|}
\hline
&{\bf G} & {\bf M} & {\bf S} & {\bf StS} \\
\hline
{\bf BLEU}& 0.36$^*$& 0.43$^*$ & 0.17 ($3 \cdot 10^{-4}$) &0.17  ($3 \cdot 10^{-4}$)   \\
\hline   
{\bf iBLEU} & 0.32$^*$ &0.40$^*$ & 0.15 ($8 \cdot 10^{-4}$) & 0.15 ($8 \cdot 10^{-4}$) \\
\hline
{\bf SARI} & -0.05 (0.2)& -0.11 (0.02)  &  0.18 ($10^{-4}$) & 0.19 ($6 \cdot 10^{-5}$)\\
\hline \hline
{\bf -LD$_{SC}$} & 0.65$^*$ & 0.66$^*$ &0.21$^*$ & 0.20 ($10^{-5}$) \\
\hline   
\end{tabular}
\hfill
\caption{\small Sentence-level Spearman correlation (and $p$-values) between the automatic metrics and the human ratings for ``HSplit as Reference Setting''. $^*p<10^{-5}$.}  
\label{tab:correlation_splitreferences}
\end{table}
\end{center}
\vspace{-1cm}

\subsection{Results with HSplit as Reference Setting}\label{sec:splitting_refs}

We turn to examining whether BLEU may be adapted to address sentence splitting, if provided with references that include splittings.

\paragraph{Description of the Human Evaluation Scores.}

In the "HSplit as Reference Reference Setting", where 6 systems are considered, the max-min difference at the system level is 0.16 ($\sigma$ = 0.06) for G, 0.37 for M ($\sigma$ = 0.15), and 0.41 for S and StS ($\sigma$ equals 0.20 and 0.19 respectively). At the sentence level, considering 420 sentences (70 for each of the systems), the G and M scores vary from 1 to 5 ($\sigma$ equals 0.99 and 0.88 respectively), and the S and StS scores from -2 to 2 ($\sigma$ equals 0.63).

\paragraph{Correlation with Human Evaluation.}

On the system-level Spearman correlation between BLEU and human judgments, we
find that while correlation with G is high (0.57, $p=0.1$), 
 it is low for M (0.11, $p= 0.4$),
and negative for S (-0.70, $p= 0.06$) and StS (-0.60, $p = 0.1$). Sentence-level correlations of BLEU and iBLEU are positive,
but they are lower than those obtained by LD$_{SC}$. See Table \ref{tab:correlation_splitreferences}. 

To recap, results in this section demonstrate that even when evaluated against references that
focus on sentence splitting, BLEU fails to capture the simplicity and structural simplicity of the output.

\section{Conclusion} \label{sec:conclusion}

In this paper we argued that BLEU is not suitable for TS evaluation, showing 
that
(1) BLEU negatively correlates with simplicity,
and that 
(2) even as a measure of grammaticality or meaning preservation it is
comparable to, or worse than -LD$_{SC}$, which requires no references.
Our findings suggest that BLEU should not be used for the evaluation of TS in general and sentence splitting in particular,
and motivate the development of alternative methods for structural TS evaluation, such as \citep{S18naacl}.

\section*{Acknowledgments}

We would like to thank the annotators for participating in our generation and evaluation experiments.
We also thank the anonymous reviewers for their helpful advices. This work was partially supported by the Intel Collaborative Research Institute for Computational Intelligence (ICRI-CI) and by the Israel Science Foundation
(grant No. 929/17), as well as by the HUJI Cyber Security Research
Center in conjunction with the Israel National Cyber
Bureau in the Prime Minister's Office.
\vspace{-0.5cm}

\bibliographystyle{acl_natbib_nourl}
\bibliography{bibliobleu}

\begin{thebibliography}{34}
\expandafter\ifx\csname natexlab\endcsname\relax\def\natexlab#1{#1}\fi

\bibitem[{Aharoni and Goldberg(2018)}]{AG18}
Roee Aharoni and Yoav Goldberg. 2018.
\newblock Split and rephrase: Better evaluation and a stronger baseline.
\newblock In \emph{Proc. of ACL'18, Short papers}, pages 719--724.
\newblock \url{http://aclweb.org/anthology/P18-2114}.

\bibitem[{Callison-Burch et~al.(2006)Callison-Burch, Osborne, and Koehn}]{C06}
Chris Callison-Burch, Miles Osborne, and Philipp Koehn. 2006.
\newblock Re-evaluating the role of {BLEU} in machine translation.
\newblock In \emph{Proc. of EACL'06}, pages 249--256.
\newblock \url{http://www.aclweb.org/anthology/E06-1032}.

\bibitem[{Chandrasekar et~al.(1996)Chandrasekar, Doran, and Srinivas}]{C96}
Raman Chandrasekar, Christine Doran, and Bangalore Srinivas. 1996.
\newblock Motivations and methods for sentence simplification.
\newblock In \emph{Proc. of COLING'96}, pages 1041--1044.
\newblock \url{http://aclweb.org/anthology/C/C96/C96-2183.pdf}.

\bibitem[{Cohen(1968)}]{C68}
Jacob Cohen. 1968.
\newblock Weighted kappa: {N}ominal scale agreement provision for scaled
  disagreement or partial credit.
\newblock \emph{Psychological bulletin}, 70(4):213.

\bibitem[{Graham(2015)}]{G15}
Yvette Graham. 2015.
\newblock Re-evaluating automatic summarization with {BLEU} and 192 shades of
  {ROUGE}.
\newblock In \emph{Proc. of EMNLP'15}, pages 128--137.
\newblock \url{http://www.aclweb.org/anthology/D15-1013}.

\bibitem[{Kincaid et~al.(1975)Kincaid, Jr., Rogers, and Chissom}]{K75}
J.~Peter Kincaid, Robert P.~Fishburne Jr., Richard~L. Rogers, and Brad~S.
  Chissom. 1975.
\newblock Derivation of new readability formulas (automated readability index,
  fog count and {F}lesch reading ease formula) for {N}avy enlisted personnel.
\newblock Technical report, Defense Technical Information Center (DTIC)
  Document.

\bibitem[{Koehn et~al.(2007)Koehn, Hoang, Birch, Callison-Buch, Federico,
  Bertoldi, Cowan, Shen, Moran, Zens, Dyer, Bojar, Constantin, and
  Herbst}]{K07}
Philipp Koehn, Hieu Hoang, Alexandra Birch, Chris Callison-Buch, Marcello
  Federico, Nicola Bertoldi, Brooke Cowan, Wade Shen, Christine Moran, Richard
  Zens, Chris Dyer, Ond\v{r}ej Bojar, Alexandra Constantin, and Evan Herbst.
  2007.
\newblock Moses: open source toolkit for statistical machine translation.
\newblock In \emph{Proc. of ACL'07 on interactive poster and demonstration
  sessions}, pages 177--180.
\newblock \url{http://aclweb.org/anthology/P/P07/P07-2045.pdf}.

\bibitem[{Koehn and Monz(2006)}]{KM06}
Philipp Koehn and Christof Monz. 2006.
\newblock Manual and automatic evaluation of machine translation between
  {E}uropean languages.
\newblock In \emph{Proc. of the Workshop on Statistical Machine Translation}.
\newblock \url{http://www.aclweb.org/anthology/W06-3114}.

\bibitem[{Li and Nenkova(2015)}]{LN15}
Junyi~Jessi Li and Ani Nenkova. 2015.
\newblock Detecting content-heavy sentences: {A} cross-language case study.
\newblock In \emph{Proc. of EMNLP'15}, pages 1271--1281.
\newblock \url{http://www.aclweb.org/anthology/D15-1148}.

\bibitem[{Loper and Bird(2002)}]{LB02}
Edward Loper and Steven Bird. 2002.
\newblock {NLTK}: the natural language toolkit.
\newblock In \emph{Proc. of EMNLP'02}, pages 63--70.
\newblock \url{http://www.aclweb.org/anthology/W/W02/W02-0109.pdf}.

\bibitem[{Ma and Sun(2017)}]{MS17}
Shuming Ma and Xu~Sun. 2017.
\newblock A semantic relevance based neural network for text summarization and
  text simplification.
\newblock ArXiv:1710.02318 [cs.CL]. \url{https://arxiv.org/pdf/1710.02318.pdf}.

\bibitem[{Mason and Kendall(1979)}]{MK79}
Jana~M. Mason and Janet~R. Kendall. 1979.
\newblock Facilitating reading comprehension through text structure
  manipulation.
\newblock \emph{Alberta Journal of Medical Psychology}, 24:68--76.

\bibitem[{Mishra et~al.(2014)Mishra, Soni, Sharma, and Sharma}]{M14}
Kshitij Mishra, Ankush Soni, Rahul Sharma, and Dipti~Misra Sharma. 2014.
\newblock Exploring the effects of sentence simplification on {H}indi to
  {E}nglish {M}achine {T}ranslation systems.
\newblock In \emph{Proc. of the Workshop on Automatic Text Simplification:
  Methods and Applications in the Multilingual Society}, pages 21--29.
\newblock \url{http://www.aclweb.org/anthology/W14-5603}.

\bibitem[{Narayan and Gardent(2014)}]{NG14}
Shashi Narayan and Claire Gardent. 2014.
\newblock Hybrid simplification using deep semantics and machine translation.
\newblock In \emph{Proc. of ACL14}, pages 435--445.
\newblock \url{http://aclweb.org/anthology/P/P14/P14-1041.pdf}.

\bibitem[{Narayan and Gardent(2016)}]{NG16}
Shashi Narayan and Claire Gardent. 2016.
\newblock Unsupervised sentence simplification using deep semantics.
\newblock In \emph{Proc. of INLG'16}, pages 111--120.
\newblock \url{http://aclweb.org/anthology/W/W16/W16-6620.pdf}.

\bibitem[{Narayan et~al.(2017)Narayan, Gardent, Cohen, and Shimorina}]{N17}
Shashi Narayan, Claire Gardent, Shay~B. Cohen, and Anastasia Shimorina. 2017.
\newblock Split and rephrase.
\newblock In \emph{Proc. of EMNLP'17}, pages 617--627.
\newblock \url{http://aclweb.org/anthology/D17-1064}.

\bibitem[{Nisioi et~al.(2017)Nisioi, {\v S}tajner, Ponzetto, and Dinu}]{Ni17}
Sergiu Nisioi, Sanja {\v S}tajner, Simone~Paolo Ponzetto, and Liviu~P. Dinu.
  2017.
\newblock Exploring neural text simplification models.
\newblock In \emph{Proc. of ACL'17 (Short paper)}, pages 85--91.
\newblock \url{http://www.aclweb.org/anthology/P17-2014}.

\bibitem[{Papineni et~al.(2002)Papineni, Roukos, Ward, and Zhu}]{P02}
Kishore Papineni, Salim Roukos, Todd Ward, and Wei-Jing Zhu. 2002.
\newblock {BLEU}: a method for automatic evaluation of machine translation.
\newblock In \emph{Proc. of ACL'02}, pages 311--318.
\newblock \url{http://aclweb.org/anthology/P/P02/P02-1040.pdf}.

\bibitem[{Park and Levy(2011)}]{PL11}
Y.~Albert Park and Roger Levy. 2011.
\newblock Automated whole sentence grammar correction using a noisy channel
  model.
\newblock In \emph{Proc.of ACL-HLT'11}, pages 934--944.
\newblock \url{http://aclweb.org/anthology/P/P11/P11-1094.pdf}.

\bibitem[{Pavlick and Tetreault(2016)}]{PT16}
Ellie Pavlick and Joel Tetreault. 2016.
\newblock An empirical analysis of formality in online communication.
\newblock \emph{TACL}, 4:61--74.
\newblock \url{http://www.aclweb.org/anthology/Q16-1005}.

\bibitem[{Siddharthan(2006)}]{S06}
Advaith Siddharthan. 2006.
\newblock Syntactic simplification and text cohesion.
\newblock \emph{Research on Language and Computation}, 4:77--109.

\bibitem[{Siddharthan and Angrosh(2014)}]{SA14}
Advaith Siddharthan and M.~A. Angrosh. 2014.
\newblock Hybrid text simplification using synchronous dependency grammars with
  hand-written and automatically harvested rules.
\newblock In \emph{Proc. of EACL'14}, pages 722--731.
\newblock \url{http://aclweb.org/anthology/E/E14/E14-1076.pdf}.

\bibitem[{{\v S}tajner et~al.(2015){\v S}tajner, Bechara, and Saggion}]{Sa15}
Sanja {\v S}tajner, Hannah Bechara, and Horacio Saggion. 2015.
\newblock A deeper exploration of the standard {PB-SMT} approach to text
  simplification and its evaluation.
\newblock In \emph{Proc. of ACL'15, Short papers}, pages 823--828.
\newblock \url{http://aclweb.org/anthology/P/P15/P15-2135.pdf}.

\bibitem[{{\v S}tajner et~al.(2014){\v S}tajner, Mitkov, and Saggion}]{S14}
Sanja {\v S}tajner, Ruslan Mitkov, and Horacio Saggion. 2014.
\newblock One step closer to automatic evaluation of text simplification
  systems.
\newblock \emph{Proc. of the 3rd Workshop on Predicting and Improving Text
  Readability for Target Reader Populations}, pages 1--10.
\newblock \url{http://www.aclweb.org/anthology/W14-1201}.

\bibitem[{Sulem et~al.(2018{\natexlab{a}})Sulem, Abend, and
  Rappoport}]{S18naacl}
Elior Sulem, Omri Abend, and Ari Rappoport. 2018{\natexlab{a}}.
\newblock Semantic structural evaluation for text simplification.
\newblock In \emph{Proc. of NAACL'18}, pages 685--696.
\newblock \url{http://aclweb.org/anthology/N18-1063}.

\bibitem[{Sulem et~al.(2018{\natexlab{b}})Sulem, Abend, and Rappoport}]{S18acl}
Elior Sulem, Omri Abend, and Ari Rappoport. 2018{\natexlab{b}}.
\newblock Simple and effective text simplification using semantic and neural
  methods.
\newblock In \emph{Proc. of ACL'18}, pages 162--173.
\newblock \url{http://aclweb.org/anthology/P18-1016}.

\bibitem[{Sun and Zhou(2012)}]{SZ12}
Hong Sun and Ming Zhou. 2012.
\newblock Joint learning of a dual {SMT} system for paraphrase generation.
\newblock In \emph{Proc. of ACL'12}, pages 38--42.
\newblock \url{http://aclweb.org/anthology/P/P12/P12-2008.pdf}.

\bibitem[{Williams et~al.(2003)Williams, Reiter, and Osman}]{W03}
Sandra Williams, Ehud Reiter, and Liesl Osman. 2003.
\newblock Experiments with discourse-level choices and readability.
\newblock In \emph{Proc. of the European Natural Language Workshop (ENLG)}.

\bibitem[{Woodsend and Lapata(2011)}]{WL11}
Kristian Woodsend and Mirella Lapata. 2011.
\newblock Learning to simplify sentences with quasi-synchronous grammar and
  integer programming.
\newblock In \emph{Proc. of EMNLP'11}, pages 409--420.
\newblock \url{http://aclweb.org/anthology/D/D11/D11-1038.pdf}.

\bibitem[{Wubben et~al.(2012)Wubben, van~den Bosch, and Krahmer}]{W12}
Sander Wubben, Antal van~den Bosch, and Emiel Krahmer. 2012.
\newblock Sentence simplification by monolingual machine translation.
\newblock In \emph{Proc. of ACL'12}, pages 1015--1024.
\newblock \url{http://aclweb.org/anthology/P/P12/P12-1107.pdf}.

\bibitem[{Xu et~al.(2016)Xu, Napoles, Pavlick, Chen, and Callison-Burch}]{Xu16}
Wei Xu, Courtney Napoles, Ellie Pavlick, Quanze Chen, and Chris Callison-Burch.
  2016.
\newblock Optimizing statistical machine translation for text simplification.
\newblock \emph{TACL}, 4:401--415.
\newblock \url{http://aclweb.org/anthology/Q/Q16/Q16-1029.pdf}.

\bibitem[{Zhang and Lapata(2017)}]{ZL17}
Xingxing Zhang and Mirella Lapata. 2017.
\newblock Sentence simplification with deep reinforcement learning.
\newblock In \emph{Proc. of EMNLP'17}, pages 595--605.
\newblock \url{http://aclweb.org/anthology/D17-1062}.

\bibitem[{Zhang et~al.(2017)Zhang, Ye, Zhao, and Yan}]{Z17}
Yaoyuan Zhang, Zhenxu Ye, Dongyan Zhao, and Rui Yan. 2017.
\newblock A constrained sequence-to-sequence neural model for sentence
  simplification.
\newblock ArXiv:1704.02312 [cs.CL]. \url{https://arxiv.org/pdf/1704.02312.pdf}.

\bibitem[{Zhu et~al.(2010)Zhu, Bernhard, and Gurevych}]{Z10}
Zhemin Zhu, Delphine Bernhard, and Iryna Gurevych. 2010.
\newblock A monolingual tree-based translation model for sentence
  simplification.
\newblock In \emph{Proc. of COLING'10}, pages 1353--1361.
\newblock \url{http://aclweb.org/anthology/C/C10/C10-1152.pdf}.

\end{thebibliography}

\end{document}